\documentclass{article}
\usepackage[dvips]{graphicx}

\begin{document}

\title{Analysis of Magnification in Depth from Defocus}
\author{Arnav Bhavsar\ \\ \ \\Dept. of Electrical Engineering\\Indian Institute of Technology Madras, India}

\date{}
\maketitle
\begin{abstract}
In depth from defocus (DFD), when images are captured with different camera parameters, a relative magnification is induced between them. Image warping is a simpler solution to account for magnification than seemingly more accurate optical approaches. This work is an investigation into the effects of magnification on the accuracy of DFD. We comment on issues regarding scaling effect on relative blur computation. We statistically analyze accountability of scale factor, commenting on the bias and efficiency of the estimator that does not consider scale. We also discuss the effect of interpolation errors on blur estimation in a warping based solution to handle magnification and carry out experimental analysis to comment on the blur estimation accuracy.
\end{abstract}

\section{Introduction}
In DFD, relative defocus blur between images is exploited as a cue for depth \cite{n1}. However, accurately estimating blur is a difficult problem. After early works \cite{n1,n2}, there are now many methods for estimating blur addressing major issues such as noise \cite{n3}, space-variance \cite{n3,n5} and the blur model \cite{n4}. 

When images are captured with different camera settings to induce a relative blur, a relative magnification is also induced between them causing a shift between the corresponding pixels in the images. Since conventional DFD does not factor in such a shift, the blur estimation can be erroneous in presence of magnification (scaling). Approaches to handle scaling can be classified as optical \cite{n6} or warping based \cite{n7,n8}. Nayar and Watanabe \cite{n6} perform an optical correction by using an additional aperture at an analytically calculated position. But this adds to hardware and cost. Ghita et al. \cite{n7} consider magnification correction in an active DFD scheme by interpolation that exploits an active illumination pattern. An interpolation based approach proposed by Darell and Wohn \cite{n8} for shape from focus computes the warping by observing the motion of a special pattern. Scaling in DFD depends only on camera parameters and since it is required that camera parameters be known, the scale factor is available apriori. Warping based methods may be preferable due to simplicity but estimation could be compromised due to interpolation effects \cite{n6}. However, a formal analysis of the magnification effect in DFD is totally lacking in literature. 

In this paper, we formally analyze the magnification effect in DFD theoretically and experimentally. Section $2$ points out important implications of magnification on the relative blur computation. Section $3$ provides a statistical analysis of scaling on blur estimation. Since the analysis implicitly assumes an image warping solution, section $4$ discusses effects of interpolation errors. Section $5$ provides experimental analysis regarding blur estimation accuracy and we conclude in section $6$. We limit ourselves to space-invariant blur since our goal is not to propose a new method but to analyze magnification as an important consideration in DFD. 

\section{Magnification in DFD}
For a thin lens model, the blur radius $R$ and depth $D$ can be related as  
\begin{equation}
R = rV\left(\frac{1}{F} -\frac{1}{V} -\frac{1}{D}\right)
\end{equation}
where $r$ is the aperture radius, $V$ is the distance between the lens and the image plane, $F$ is the focal length. Varying $r$ and $F$ keeping $V$ constant will cause relative blurring without inducing magnification. However, since $F$ is a function of the physical parameters of lens, changing $F$ means physically changing the lens. As mentioned in \cite{n6}, the sensitivity of blur due to change in $r$ is low. Hence a better way to induce relative blur is to vary $V$. However, this introduces a relative magnification between the two images. The scaling factor can be derived as $s =\frac{V_1}{V_2}$, where $V_1$ and $V_2$ are two instances of $V$ while capturing the two images (Fig 1). 
\begin{figure}[h]
\centering
\includegraphics[width=150pt]{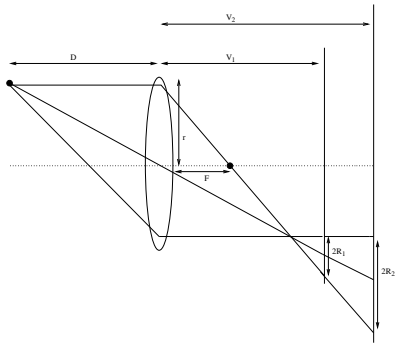}
\caption{Magnification and blurring with variation in camera parameters}
\end{figure}
Although the above scaling factor is defined for a thin lens, our intention is to note that scale is a function of known camera parameters. Hence, even in compound lenses, the scale factor can be known but with more involved calculations. Thus the amount of magnification can be known apriori. We now discuss two important implications of relative scaling on estimating blur. 
\subsection{Order of scaling and blurring}
Due to magnification, two types of transformations are involved that relate the two images viz. scaling and blurring. The order of the transformations is vital when estimating the blur. Typically, the image formation is expressed as,   
\begin{equation}
g(x) = h(x)\ast f(x) + \eta(x)
\end{equation}
In frequency domain, we have 
\begin{equation}
G(\omega) = H(\omega)F(\omega) + N(\omega)
\end{equation}
In DFD, the Gaussian point spread function (PSF) is popular as a blur model because the effects of blurring, diffraction and sampling can be expressed as a Gaussian in view of the central limit theorem \cite{n1}. The standard deviation $\sigma$ of the Gaussian PSF referred to as `blur parameter' is related to $R$ in equation (1) as $\sigma = \rho R$ where $\rho$ is a constant. Thus $H(\omega) = \exp^{-\frac{\sigma^2\omega^2}{2}}$. 

Considering the order of blurring followed by scaling, equation (3) becomes, 
\begin{equation}
G(\omega) = \frac{1}{s}H\left(\frac{\omega}{s}\right)F\left(\frac{\omega}{s}\right) + N(\omega)
\end{equation}
Hence, $H(\frac{\omega}{s}) = \exp^{-\frac{\sigma^2\omega^2}{2s^2}}$. This effectively means that scaling of the image $f$ is followed by blurring with blur parameter $\frac{\sigma}{s}$. However referring to Figure 1 and its related discussion, we see that the shift in the pixels is due to scaling $s$ and the blur around these corresponding pixel is given by equation (1). According to the above thin lens model, these shifted points are spread by a blur having parameter $\sigma$ rather than $\frac{\sigma}{s}$. Thus we conclude that the order of scaling followed by blurring is more valid than vice-versa. Thus the image formation when considering scaling should be
\begin{equation}
G(\omega) = \frac{1}{s}H(\omega)F\left(\frac{\omega}{s}\right) + N(\omega)
\end{equation}

\subsection{Effect of scaling on relative blur computation}
Generally, given two blurred images $g_1$ and $g_2$ a model that is often used \cite{n9} is
\begin{equation}
g_2(x) = h_r(x)\ast g_1(x) + \eta(x)
\end{equation}
where $\ast$ denotes convolution. This can be derived as follows 
\begin{eqnarray}
G_i(\omega) &=& H_i(\omega)F(\omega)\hspace{1cm} i = 1,2\\
\Rightarrow\frac{G_2(\omega)}{G_1(\omega)} &=& \frac{H_2(\omega)}{H_1(\omega)} = H_r(\omega)\nonumber\\
\Rightarrow G_2(\omega) &=& H_r(\omega)G_1(\omega)\nonumber
\end{eqnarray}
In spatial domain adding noise this gives equation (6). The blur parameter of the relative blur $h_r(x)$ (or $H_r(\omega)$) is $\sqrt{\sigma_2^2 - \sigma_1^2}$ where $\sigma_1$ and $\sigma_2$ are the blur parameters of $H_1(\omega)$ and $H_2(\omega)$. This can be deduced from the Gaussian PSF of the blur and the fact that $H_r(\omega) = \frac{H_2(\omega)}{H_1(\omega)}$. Estimating blur from equation (6) actually means solving for $\sqrt{\sigma_2^2 - \sigma_1^2}$. This relationship along with $\sigma_2 = \alpha\sigma_1 + \beta$, can be used to solve for $\sigma_1$ or $\sigma_2$. Here $\alpha$ and $\beta$ are dependent on camera parameters as $\alpha = \frac{r_1V_1}{r_2V_2}$ and $\beta = \rho r_1V_1\left(\frac{1}{F_1} -\frac{1}{V_1} -\frac{1}{F_2} + \frac{1}{V_2}\right)$ in a general case of varying all the camera parameters \cite{n2}. Considering scale in the frequency domain version of equation (6) results in a simple extension of (5) as 
\begin{equation}
G_2(\omega) = \frac{1}{s}H_r(\omega)G_1\left(\frac{\omega}{s}\right) + N(\omega)
\end{equation}
Apparently, this should resulting in the estimation of $\sqrt{\sigma_2^2 - \sigma_1^2}$, the blur parameter of $H_r(\omega)$. However such an extension of (5) is incorrect as shown below. Actually, the scale consideration should be in the basic image formation i.e. equation (7),  
\begin{eqnarray}
G_i(\omega) &=& \frac{1}{s_i}H_i(\omega)F\left(\frac{\omega}{s_i}\right)\hspace{1cm}i = 1,2\\
\Rightarrow G_2(\omega) &=& \frac{s_1}{s_2}\frac{H_2(\omega)}{H_1\left(\frac{s_1}{s_2}\omega\right)}{G_1\left(\frac{s_1}{s_2}\omega\right)}\nonumber\\ 
\end{eqnarray}
Here, $s_1$ and $s_2$ are the scaling factors when the focused image  $f$ is transformed into $g_1$ and $g_2$ respectively. We note that since we do not have the focused image, we also do not have $s_1$ and $s_2$. However, we only need the relative scale $s = \frac{s_2}{s_1}$ which is known apriori from the camera parameters. We now write the modified version of equation (6) as,
\begin{equation}
g_2(x) = h_r(x)\ast g_2(sx) + \eta(x)
\end{equation}
Here, the correct relative blur parameter of $h_r$ is $\sqrt{\sigma_2^2 - s^2\sigma_1^2}$ rather than $\sqrt{\sigma_2^2 - \sigma_1^2}$. Thus when accounting for scaling one must consider this modification in the relative blur expression to solve for $\sigma_1$ or $\sigma_2$.

\section{Analysis of the scaling effect}
In this section we answer the following question. `Given two images with relative scaling and blurring, how important is scale consideration for blur estimation ?' We show that the Least-Squares (LS) estimator that does not account for scaling is both biased and inefficient. For simplicity, we assume that one of the two images is the focused image.  
\subsection{Bias in blur estimation}
From the discussion in the previous section, the true image formation model can be written as
\begin{equation}
g(x) = h(x)\ast f(sx) + \eta(x) \hspace{0.3cm} \mbox{or} \hspace{0.3cm}\underline{g} = F_s\underline{h} + \underline{\eta}
\end{equation}
where $F_s$ is a block Toeplitz image matrix corresponding to the focused image $f(sx)$, $\underline{g}$ is the observed image, $\underline{h}$ is a blur vector and $\underline{\eta}$ is a zero mean additive white Gaussian noise (AWGN) vector. A least-squares solution of the blur estimate $\hat{h}$ will be
\begin{equation}
\underline{\hat{h}} = (F_s^{T}F_s)^{-1}F_s^{T}\underline{g}
\end{equation}
The bias in this estimate is then\\ 
\begin{eqnarray}
\underline{b_s} = E(\underline{\hat{h}})-\underline{h} & = & E((F_s^{T}F_s)^{-1}F_s^{T}\underline{g}) - \underline{h} = (F_s^{T}F_s)^{-1}F_s^{T}F_s\underline{h} - \underline{h} = 0
\end{eqnarray}
However, if scaling is not taken into account during estimation, then we obtain the estimator
\begin{equation}
\underline{\hat{h}} = (F^{T}F)^{-1}F^{T}\underline{g}
\end{equation}
Since in actuality, $E(\underline{g}) = F_s\underline{h}$, this results in a non zero bias if one does not account for the scale factor $s$.
\begin{eqnarray}
\underline{b} = E(\underline{\hat{h}})-\underline{h} & = & E((F^{T}F)^{-1}F^{T}\underline{g}) - \underline{h} = (F^{T}F)^{-1}F^{T}F_s\underline{h} - \underline{h} \neq 0
\end{eqnarray}
The above bias can be explained as follows. The matrix $F$ consists of entries from the focused image $f(\textbf{x})$. The matrix $F_s$ consists of entries from $f(s\textbf{x})$, the scaled version of the focused image. To account for scale, one must convert $f(\textbf{x})$ to $f(s\textbf{x})$. Doing this will ideally make the bias to be $(F_s^{T}F_s)^{-1}F_s^{T}F_s\underline{h} - \underline{h} = 0$. Not accounting for scale will mean that we are using the focused image $f(\textbf{x})$ for estimating the blur when the actual focused image is $f(s\textbf{x})$, the scaled version of the $f(\textbf{x})$, thus inducing a bias.  

\subsection{Efficiency of the estimator}
Here we comment on the efficiency of the estimator that does and does not consider scaling. The efficiency is in the sense of the Cramer-Rao lower bound (CRLB). We do not provide the details due to space constraints. Considering the true image formation (equation (12)), the CRLB for the blur estimate $\hat{\underline{h}}$ can be shown to be
\begin{equation}
Cov(\hat{\underline{h}}) \geq  \sigma_v^2(F_s^TF_s)^{-1}
\end{equation}      
where $Cov(\hat{\underline{h}})$ is the covariance matrix of $\hat{\underline{h}}$ and $\sigma_v^2$ is the noise variance. The LS estimator $\underline{\hat{h}} = (F_s^{T}F_s)^{-1}F_s^{T}\underline{g}$ is an efficient estimator that meets the CRLB. The estimator that does not use the scaled image is $\underline{\hat{h}} = (F^{T}F)^{-1}F^{T}\underline{g}$. As shown above, this estimator incurs a bias $\underline{b}$. Suppose a new estimator which is formed by subtracting the bias $\underline{b}$ from $\underline{\hat{h}} = (F^{T}F)^{-1}F^{T}\underline{g}$ i.e. $\underline{\hat{h}_{new}} =  (F^{T}F)^{-1}F^{T}\underline{g}$ - $\underline{b}$. This new estimator will be unbiased for obvious reasons. However, covariance matrix of this estimator turns out to be, 
\begin{equation}
Cov(\underline{\hat{h}_{new}}) \geq  \sigma_v^2(F^TF)^{-1}
\end{equation}      
Since this is not equal to the CRLB, it is an inefficient estimator. Ideally, to achieve an unbiased and efficient estimate of blur parameter $\sigma$, we must utilize the scale factor $s$ to transform the observed focused image $f(x)$ to its scaled version $f(sx)$ and then compute $\hat{\sigma}$ by solving equation (12) in the least-squares sense. However, in practice, the transformation of $f(x)$ to $f(sx)$ involves image interpolation. In the next section we comment on the effects of interpolation errors on the blur estimate.

\section{Interpolation errors}
Till now, we implicitly assumed ideal image warping to alleviate the magnification effect. i.e. the images are exactly aligned after warping. However, due to interpolation this will not be so. The difference between the ideal scaled image $f_{is}(x)$ and interpolated version $f_s(x)$ is what we call `interpolation noise',
\begin{equation}
f_s(x) = f_{is}(x) + \eta_i(x)
\end{equation}      
The samples of $\eta_i(x)$ are correlated as explained next. In Figure 2 pixels $P_i$s and $M_i$s belong to the reference image and warped image, respectively. Considering bilinear interpolation as an example, the pixels $P_2, P_5$ contribute to both $M_1$ and $M_2$. Such common contributions during interpolation induce correlation in $\eta_i(x)$. Also, from the histograms in Figure 3, we can empirically deduce that the pdf of $\eta(x)$ is heavy tailed and may not be well approximated as Gaussian. Thus interpolation errors induces correlation and non-Gaussianity. From equations (12), (19) and considering ideal scaling in equation (12) 
\begin{equation}
g(x) = h(x)\ast f_{is}(x) + \eta(x) = h(x)\ast f_s(x) - h(x)\ast\eta_i(x) + \eta(x)
\end{equation}      
where $h(x)\ast\eta_i(x)$ is the correlated, non-Gaussian component of the distribution. The LS estimator considered in the previous section is efficient for AWGN \cite{n10}, and not when the distribution possesses non-Gaussianity and uncorrelatedness \cite{n11,n12}. However a class of robust M-estimators are shown to be asymptotically efficient even under such pathologies \cite{n12,n13}. We have explored the performance of some M-estimators for blur estimation. In the next section, we carry out experiments and analyze the blur estimation accuracy.
\begin{figure}[h]
\centering
\includegraphics[width=60pt]{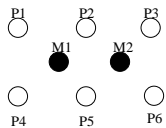}
\caption{Pixel positions in the reference image ($P_i$s) and interpolated image ($M_i$s)} 
\end{figure}
\vspace{-1cm}
\begin{figure}[h]
\centering
\begin{tabular}{c c c}
\includegraphics[width=80pt]{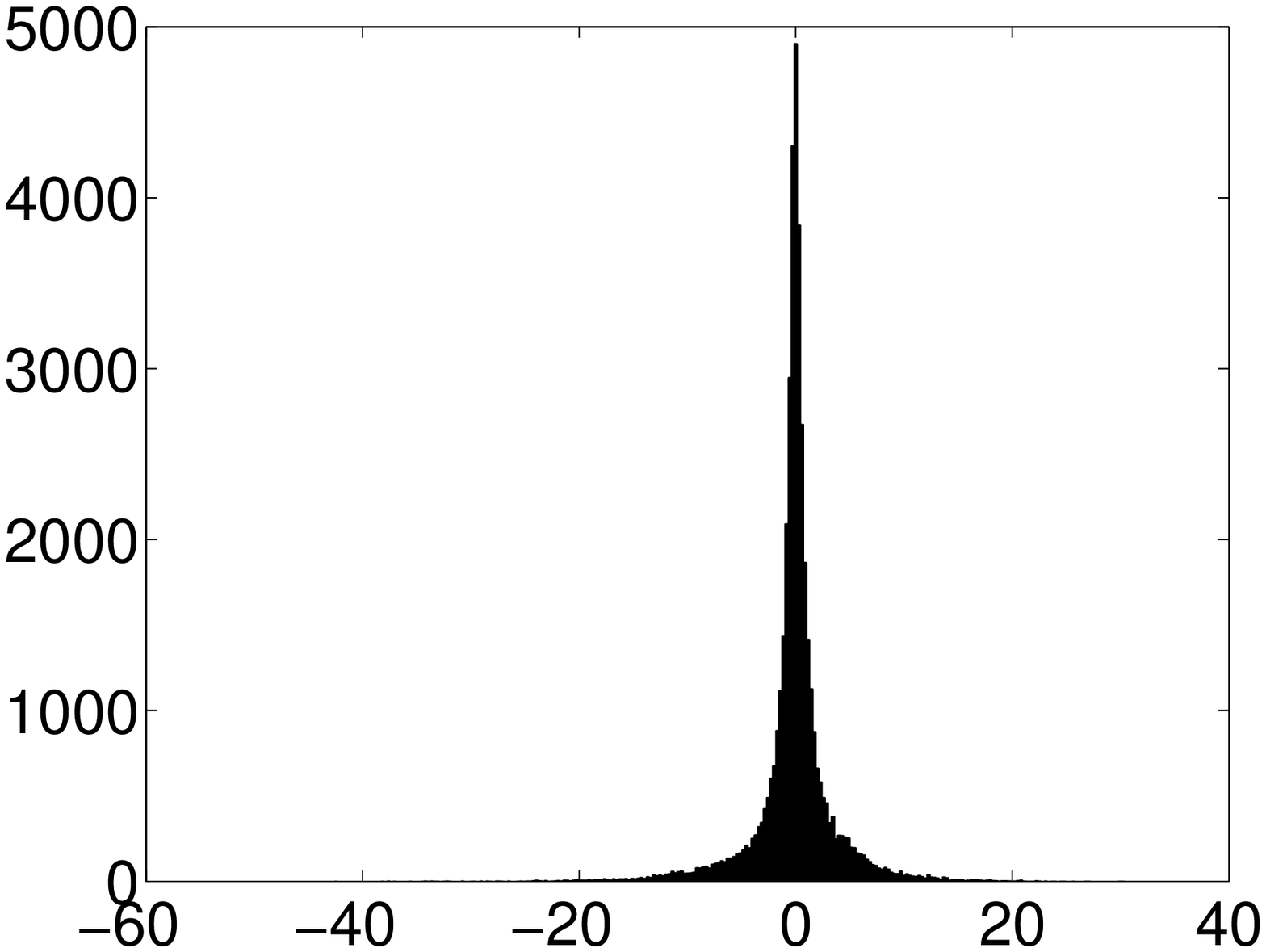} & \includegraphics[width=80pt]{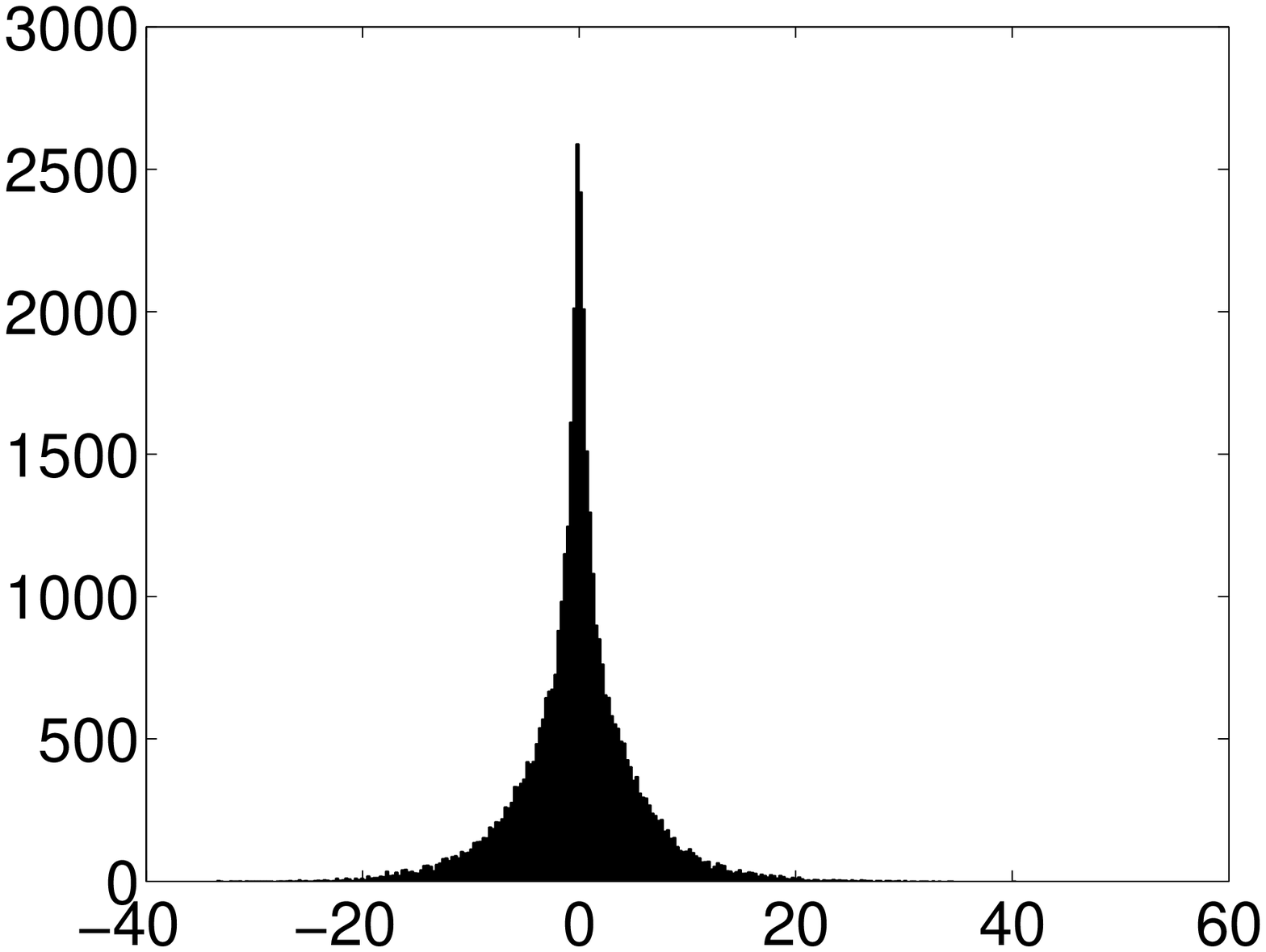} & \includegraphics[width=80pt]{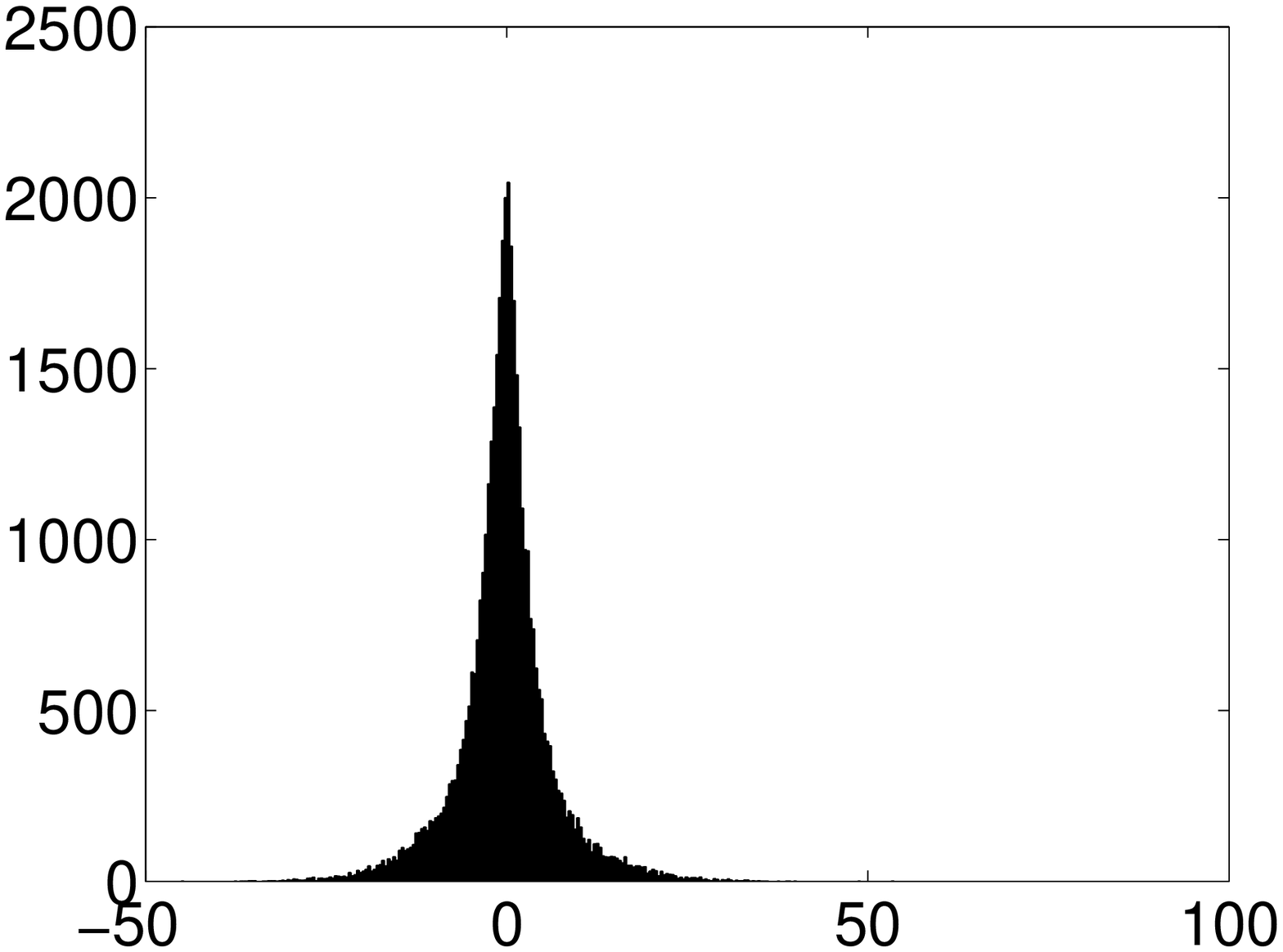}
\end{tabular}
\caption{Histograms of $\eta_i(x)$ for three images}
\end{figure}

\section{Results}
We experimentally analyzed blur estimation accuracy versus variation in scale, blur and noise.  We include two discontinuity adaptive functions and an absolute difference as robust estimators in our experiments. Discontinuity adaptive functions can serve as robust estimators due to the analogy between discontinuities and outliers \cite{n14}. Thus the estimators that we experiment with are the least squares estimator (LSE), absolute difference estimator (ABS) and discontinuity adaptive functions (DA1 and DA2)
\begin{eqnarray}
\mbox{LSE:}\hspace{0.1cm}\arg\min_{\sigma}E(x) &=& \arg\min_{\sigma}(g_1(x) - h_r(x)\ast g_2(sx))^2\\
\mbox{ABS:}\hspace{0.1cm}\arg\min_{\sigma}E(x) &=& \arg\min_{\sigma}|g_1(x) - h_r(x)\ast g_2(sx)|\nonumber\\
\mbox{DA1:}\hspace{0.1cm}\arg\min_{\sigma}E(x) &=& \arg\min_{\sigma}(1-\exp^{-(g_1(x) - h_r(x)\ast g_2(sx))^2})\nonumber\\
\mbox{DA2:}\hspace{0.1cm}\arg\min_{\sigma}E(x) &=& \arg\min_{\sigma}\left(1-\frac{1}{1+(g_1(x) - h_r(x)\ast g_2(sx))^2}\right)\nonumber
\end{eqnarray}      
%\vspace{-0.8cm} 
\textbf{Variation in scale factor:}
We vary the scale factor from 0.7 to 0.95 keeping the blur constant. The blur estimates were computed for various constants values of $\sigma_1$ and $\sigma_2$. Figure 4(a) shows the results for $\sigma_1=0.7$ and $\sigma_2=1.2$. We observe that there is no proper behavior of the estimated blur with scale variation. This is because the interpolation errors do not depend on the magnitude of the scale but rather on the shifts of individual pixels. We observe that ABS and DA estimators are very accurate as compared to the Least squares.\\\\
\textbf{Variation in blur:} This experiment involves variation of blur with a constant scale factor. In figure 4(b) we show the results for scale factor of 0.9 with $\sigma_1$ constant at 0.7 and $\sigma_2$ varying from 0.7 to 1.5. It is quite clear that generally the inaccuracy increases with blurring. However, again the inaccuracy is quite negligible for ABS, DA1 and DA2 estimators.\\\\
\textbf{Variation in noise:} We experimented with noise variances from 1 to 25 with constant scale and blur. Fig 4(c) shows results for $s = 0.9$, $\sigma_1 = 1$, $\sigma_2 = 1.5$.  For the LSE, the error is fairly constant but large. For the ABS estimator, the error increases with noise but it is quite small. The DA estimators incur negligible error. 
%\vspace{-0.3cm}
\begin{figure}[h]
\centering
\begin{tabular}{c c c}
\hspace{-1cm}
\includegraphics[width=130pt]{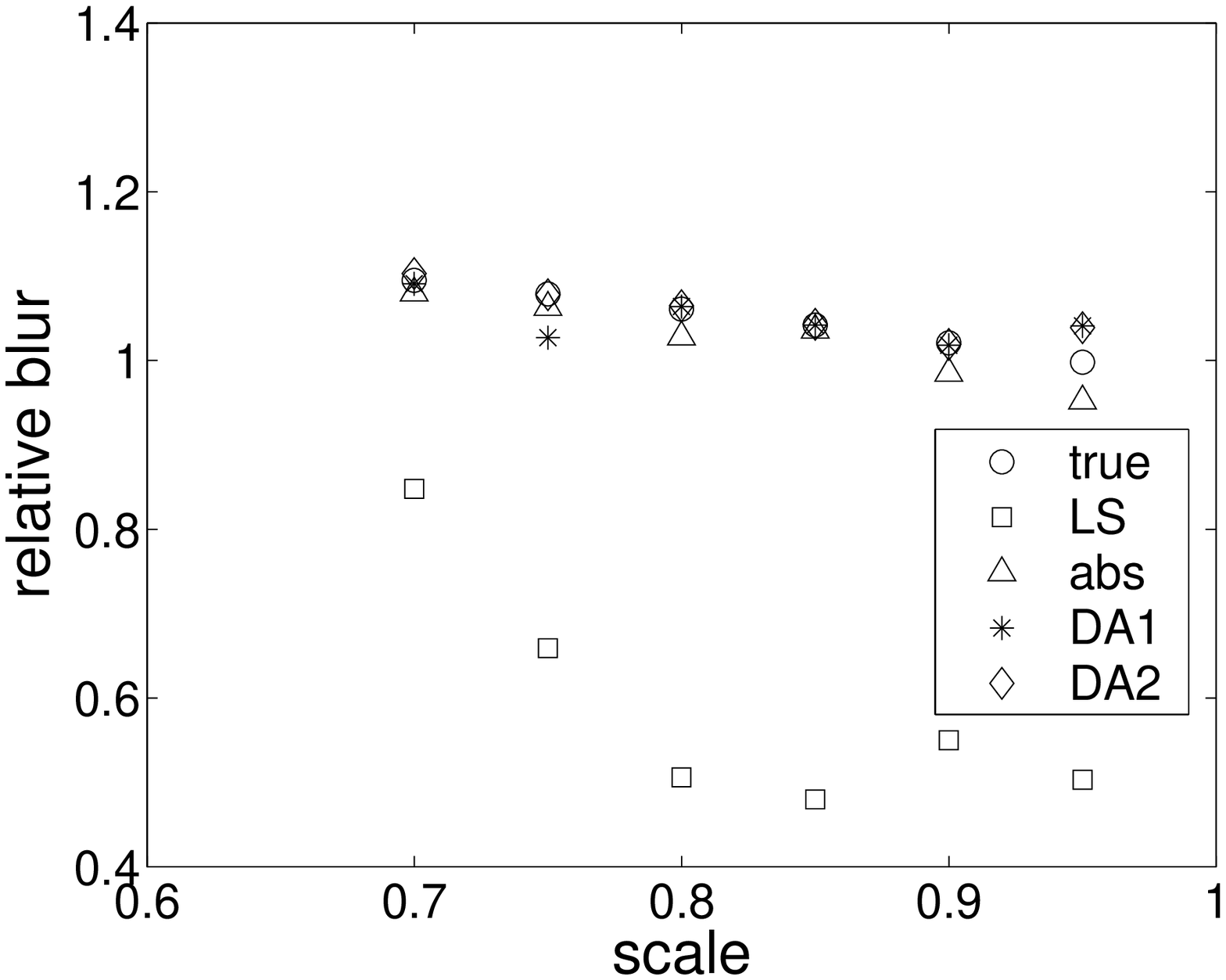} & \includegraphics[width=130pt]{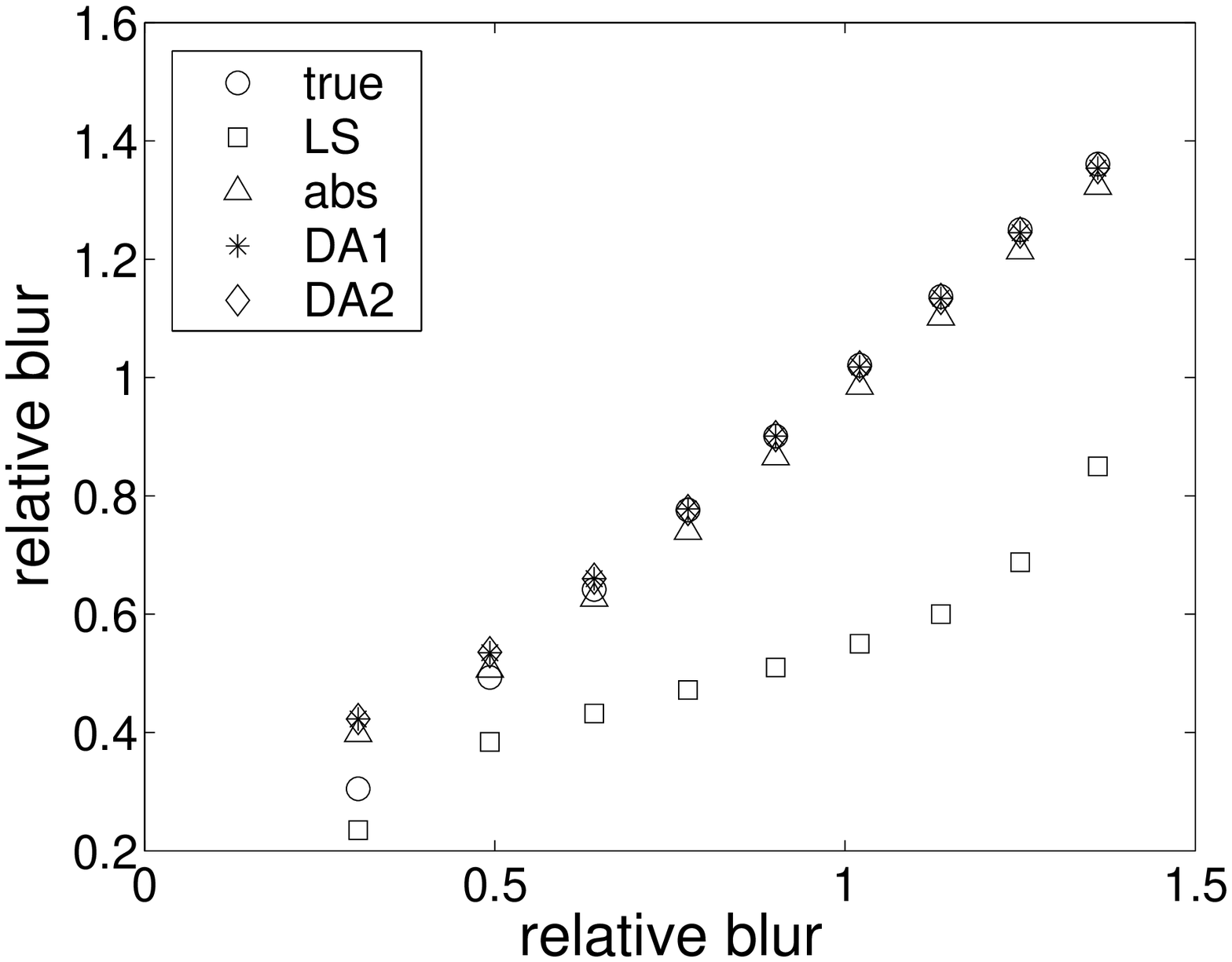} & \includegraphics[width=130pt]{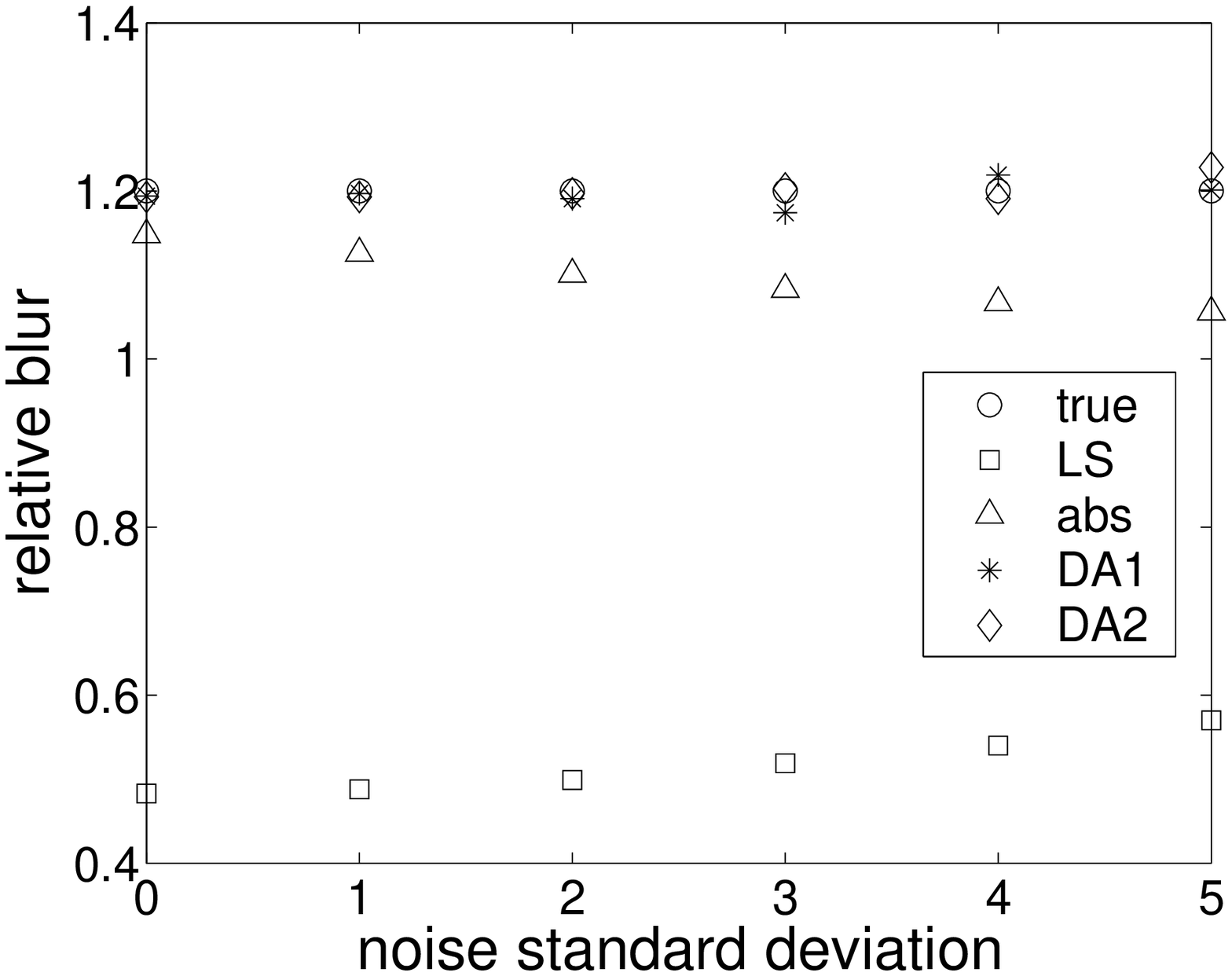}\\(a) & (b) & (c)
\end{tabular}
\caption{Blur estimation results (a) Scale variation (b) Blur variation (c) Noise variation}
\end{figure}
%\vspace{-1cm}

Thus, estimation accuracy in warping based solution depends lot on the estimator. The LS estimator is inaccurate as interpolation errors cause violation of the underlying assumptions concerning the pdf. However, robust estimators are very accurate despite interpolation effects.

\section{Conclusion}
This work analyzed the inherent magnification effect in DFD. Important issues such as the order of scaling - blurring and the effect of scaling on relative blur computation were discussed. We then carried out statistical analysis and concluded that the estimator that does not handle scaling is both biased and inefficient. Since an image warping solution was inherently assumed, we scrutinized the effect of interpolation on blur estimation accuracy. We conclude that blur estimation using robust estimators performs very well.

\end{document}